\documentclass[runningheads]{llncs}
\usepackage{graphicx}
\usepackage{amsmath}
\usepackage{cite}
\usepackage{algorithmic}
\usepackage{algorithm}
\usepackage{array}
\usepackage{textcomp}
\usepackage{stfloats}
\usepackage{url}
\usepackage{verbatim}
\usepackage{color}
\usepackage{booktabs}
\usepackage{xcolor}
\usepackage{multirow}
\usepackage{enumitem}
\definecolor{citecolor}{HTML}{0071bc} 
\usepackage[colorlinks, linkcolor=red,  anchorcolor=blue, citecolor=citecolor]{hyperref}
\usepackage{float}
\definecolor{SeaGreen4}{RGB}{0,205,102} 
\definecolor{SlateBlue}{RGB}{106,90,205} 
\definecolor{DarkRed}{RGB}{178,34,34} 
\usepackage{times}
\usepackage{soul}
\usepackage[switch]{lineno}
\usepackage{amssymb}
\usepackage{pifont}
\usepackage{makecell}
\usepackage{colortbl}
\definecolor{mygray}{gray}{.9}
\definecolor{mypink}{rgb}{.99,.91,.95}
\definecolor{mycyan}{cmyk}{.3,0,0,0}
\usepackage[misc]{ifsym}

\usepackage{subcaption}  

\begin{document}

\title{Learning Bottleneck Transformer for Event Image-Voxel Feature Fusion based Classification
\thanks{Corresponding author: Lan Chen, Email: (\email{chenlan@ahu.edu.cn})}}  



%
%

\author{Chengguo Yuan\inst{1} \and
Yu Jin\inst{1} \and 
Zongzhen Wu\inst{1} \and 
Fanting Wei\inst{1} \and 
Yangzirui Wang\inst{1} \and \\ 
Lan Chen (\Letter)\inst{1} \and 
Xiao Wang\inst{1}}
\authorrunning{Chengguo Yuan et al.}
%
\institute{1. Anhui University, Hefei City, 230601, Anhui Province, China \\ \email{\{jy\_0x4f, 17398389386\}@163.com}, \email{\{e21301283, E02114335, E02114336\}@stu.ahu.edu.cn}, \email{\{chenlan, xiaowang\}@ahu.edu.cn} 
}
\maketitle              

\begin{abstract}
Recognizing target objects using an event-based camera draws more and more attention in recent years. Existing works usually represent the event streams into point-cloud, voxel, image, etc, and learn the feature representations using various deep neural networks. Their final results may be limited by the following factors: monotonous modal expressions and the design of the network structure. To address the aforementioned challenges, this paper proposes a novel dual-stream framework for event representation, extraction, and fusion. This framework simultaneously models two common representations: event images and event voxels. By utilizing Transformer and Structured Graph Neural Network (GNN) architectures, spatial information and three-dimensional stereo information can be learned separately. Additionally, a bottleneck Transformer is introduced to facilitate the fusion of the dual-stream information. Extensive experiments demonstrate that our proposed framework achieves state-of-the-art performance on two widely used event-based classification datasets. 
The source code of this work is available at: \url{https://github.com/Event-AHU/EFV_event_classification} 
\end{abstract}

\section{Introduction}

Recognizing the category of a given object is a fundamental problem in computer vision. Most of the previous classification models are developed for frame-based cameras, in other words, these recognition models focus on encoding and learning the representation of RGB frames. With the rapid development of deep learning, frame-based classification achieves significant improvement in recent years. Representative deep models (e.g., the AlexNet~\cite{ul2018alexnet}, ResNet~\cite{he2016deep}, and Transformer~\cite{vaswani2017attention}) and datasets (e.g., ImageNet~\cite{krizhevsky2017imagenet}) are proposed one after another. However, the recognition performance in challenging scenarios is still far from unsatisfactory, including heavy occlusion, fast motion, and low illumination.

\begin{figure*}
\center
\includegraphics[width=\textwidth]{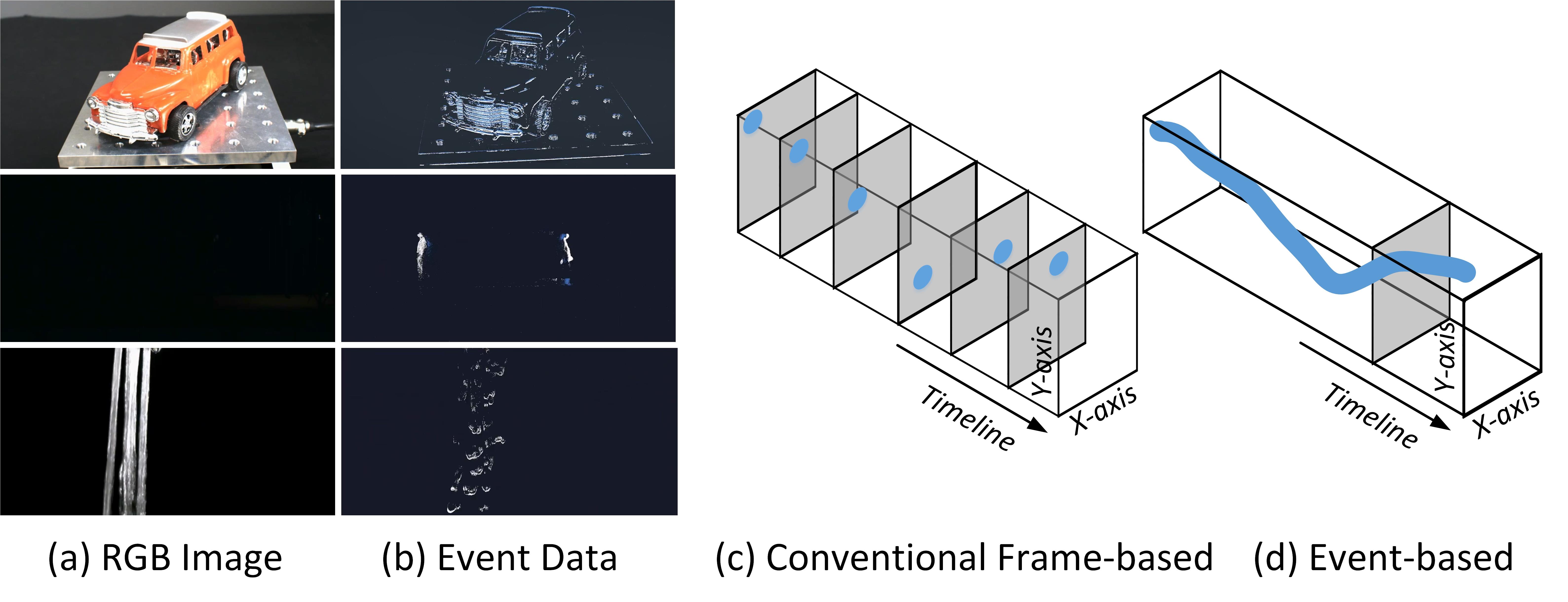}
\caption{Comparison of the frame- and event-based cameras~\url{https://youtu.be/6xOmo7Ikwzk}. (a, b) shows representative samples in regular scenarios, low-illumination, and fast motion. (c, d) illustrates the different types of raw data representation of frame- and event-based cameras. } 
\label{FrameEventCompare}
\end{figure*}

To improve object recognition in challenging scenarios, some researchers have started leveraging other sensors to obtain more effective signal inputs, thus enhancing recognition performance~\cite{9795869}. Among them, one of the most representative sensors is the event camera, also known as DVS (Dynamic Vision Sensor), which has been widely exploited in computer vision~\cite{wang2021visevent, tang2022coesot, zhu2022eventPSNN}. This paper focuses on using event cameras for object recognition. As shown in Fig.~\ref{FrameEventCompare}, different from the frame-based camera which records the light intensity for each pixel simultaneously, the event camera captures pulse signals asynchronously based on changes in light intensity, recording binary digital values of either zero or one. Typically, an increase in brightness is denoted as an ON event, while a decrease corresponds to an OFF event. An event pulse signal can be represented as a quadruple ($x, y, t, p$), where $x, y$ represents the spatial position information, $t$ represents the timestamp, and $p$ represents the polarity, i.e., ON/OFF event. Many works demonstrate that the event camera performs better in High Dynamic Range (HDR), high temporal resolution, low latency response, and strong robustness. Therefore, utilizing event cameras for object recognition is a research direction that holds great research value and practical potential.

Recently, researchers have already conducted studies on object recognition using event cameras and have proposed various approaches to address this task, including CNN (Convolutional Neural Network)~\cite{wang2019evGait}, GNN (Graph Neural Network)~\cite{wang2021eventGNN}, Transformer~\cite{vaswani2017attention}, etc. 
Although these methods have achieved good accuracy by representing and learning events from different perspectives, they are still limited by the following aspects: 
\textbf{Firstly}, they rely on a single event representation form, such as images, point clouds, or voxels, which may limit the expressiveness and versatility of the learned features. Different event representation forms may capture different aspects of the data, and using only one representation may lead to the loss of valuable information.
\textbf{Secondly}, the current methods are constrained to using only one of the deep learning architectures, such as CNNs, GNNs, or Transformers, for feature learning. Each architecture has its strengths and limitations in capturing different types of patterns and dependencies in data. By restricting the choice to a single architecture, the methods may not fully exploit the potential benefits and complementary strengths of different architectures.
To address these limitations, future research should explore approaches that can integrate multiple event representation forms and leverage the combined power of different deep learning architectures. This could involve developing novel fusion techniques or hybrid architectures that can effectively capture and leverage diverse features and dependencies present in event data. By doing so, we can potentially enhance the performance and flexibility of event-based object recognition methods.

To address the aforementioned issues, in this work, we propose an effective dual-stream event information processing framework, referred to as EFV, as shown in Fig.~\ref{framework}. Specifically, we first transform the dense event point cloud signals into event images and event voxel representations. For the input of image frames, we utilize advanced spatiotemporal Transformer networks to learn spatiotemporal features. For voxel input, considering the sparsity of events, we employ a top-k selection method to sample meaningful signals for constructing a structured graph, and then use GNN (Graph Neural Network) to learn these volumetric structured features. Importantly, we introduce the Bottleneck Transformer to integrate these two types of feature representations, which are ultimately input to the dense layer for classification. It is easy to find that our proposed EFV possesses the characteristics of efficient event information processing, integration of multiple feature representations, spatiotemporal modeling capability, consideration of event sparsity, and accurate classification capability.

To sum up, the main contributions of this work can be concluded as the following two aspects: 

$\bullet$ We propose an effective framework for recognition in event-based cameras, utilizing Event Image-Voxel feature representation and fusion. 

$\bullet$ The introduction of the Bottleneck Transformer enables the interaction and fusion of dual-stream information, leading to improved recognition results.

\section{Related Work} 
In this section, we give an introduction to Event-based Recognition\footnote{\url{https://github.com/Event-AHU/Event_Camera_in_Top_Conference}}, Graph Neural Networks, and Bottleneck Transformer.

\noindent 
\textbf{Event-based Recognition. } 
Current research on event-based recognition can be divided into three distinct streams: CNN-based~\cite{wang2019ev}, SNN (Spiking Neural Networks)-based~\cite{fang2020exploiting, fang2021incorporating}, and GNN-based models~\cite{bi2019graph, bi2020graph, wang2021event}. For the CNN-based models, Wang et al.~\cite{wang2019ev} proposed an event-based gait recognition (EV-gait) method, which effectively removes noise via motion consistency. SNN is also utilized for encoding the event stream in order to achieve energy-efficient recognition. A kind of highly efficient conversion of ANN to SNN method is put forward by Peter and others~\cite{diehl2015fast}, the method involves the balance of the weights and thresholds, while achieving lower latency and requiring fewer operations. In~\cite{perez2021sparse}, a sparse backpropagation method for SNN was introduced by redefining the surrogate gradient function form. Fang et al.\cite{fang2021deep} propose spike element-wise (SEW) ResNet to implement residual learning for deep SNNS, while proving that SEW ResNet can easily implement identity mapping and overcome the vanishing/exploding gradient problem of Spiking ResNet. 
Wang et al. propose a hybrid SNN-ANN framework for RGB-Event based recognition by fusing the memory support Transformer and spiking neural networks, termed SSTFormer~\cite{wang2023sstformer}. 
Jiang et al. propose to aggregate the event point and voxel using absorbing graph neural networks for event-based recognition~\cite{jiang2023pointvoxel}.  

For point cloud based representation, Wang et al.~\cite{wang2019space} treat the event stream as a set of 3D points in space-time, i.e., space-time event clouds, and adopt the PointNet~\cite{qi2017pointnet} architecture, which directly takes the point cloud as input and outputs the class label for the entire input or each point segment/part label for each input point. Xie et al. \cite{xie2022vmv} propose VMV-GCN, a voxel-wise graph learning model designed to integrate multi-view volumetric data. Li et al. \cite{li2022event} introduce a Transformer network to directly process event sequences in its native vector tensor format to effectively represent the temporal and spatial correlations of input raw events, thereby generating effective spatio-temporal features for the task. Different from previous works, this paper designs an event recognition method based on Transformer and graph convolutional neural network, which transmits bimodal information through a specific method and learns a unified feature representation, so as to represent event data more effectively.


\noindent 
\textbf{Graph Neural Networks. } 
One notable application of GNNs in event data recognition is gait recognition. Wang et al. \cite{wang2021event} propose a 3D graph neural network specifically designed for gait recognition. The model leverages the graph structure to capture the spatial and temporal dependencies in gait patterns. 
Bi et al. introduce the concepts of residual Graph Convolutional Neural Networks (RG-CNN) and Graph2Grid blocks \cite{bi2020graph}, \cite{bi2019graph}, which exploit graph structure to extract and exploit spatial and temporal information from event data. 
The Asynchronous, Event-based Graph Neural Networks (AEGNN) proposed by \cite{schaefer2022aegnn} addresses the processing of events as “evolving” spatio-temporal graphs. 
In the field of object recognition, Li et al. \cite{li2021graph} introduce SlideGCN, a GNN-based model that focuses on fast graph construction using a radius search algorithm. 
Different from previous works, we adopt a graph Convolutional Neural Network (GCN) to process the graph data and connect the outputs of GCN and ST-Transformer module for accurate event-based pattern recognition.

\noindent 
\textbf{Bottleneck Transformer. }  
The traditional Transformer model has the problem of large computation and memory overhead when processing large-size images. 
Srinivas et al. \cite{srinivas2021bottleneck} propose a novel network architecture called Bottleneck Transformer, which achieves dimensionality reduction of spatial attention by introducing a "bottleneck layer" between high-resolution and low-resolution representations of input features, thereby reducing computational consumption and increasing model scalability. 
Li et al. \cite{li2019enhancing} introduce a local multi-head self-attention mechanism and a novel position encoding method to solve the scalability bottleneck of Transformer under GPU memory constraints. 
Nagrani et al. \cite{nagrani2021attention} propose a multimodal bottleneck converter (MBT) and guided the bottlenecks in it to connect across modes. 
Song et al.\cite{song2022bs2t} propose a new model BS2T that captures long-range dependencies between pixels in HS images by leveraging the self-attention mechanism in Transformers. 
In addition, we innovatively introduce bottleneck Transformer to promote the fusion of dual-stream information and improve the performance of module fusion.

\section{Our Proposed Approach} 

\subsection{Overview} 
Given an input event stream consisting of hundreds of thousands of events, our approach involves several steps to enhance the representation. Initially, we employ event frame stacking and voxel construction techniques to generate event frame and voxel representations, respectively. Subsequently, we utilize two intermediate representations, namely event frame and voxel graph, to capture the spatio-temporal relationships within the event stream. To further improve the feature descriptors for event frame and graph-based event representation, we propose a novel dual branch learning network. Finally, we combine these representations to create a comprehensive representation for event data, enabling effective recognition. The overall framework is depicted in Fig.~\ref{framework}. In the following sections, we provide a detailed explanation of each module.

\subsection{Network Architecture} 

\noindent 
\textbf{Input Representation. } 
Considering the large amount of data and computational complexity, it is necessary to employ some down-sampling techniques to reduce the number of events. In this paper, we adopt two kinds of sampling techniques to obtain the compressed event representations. We first transform the asynchronous event flow into the synchronous event images by stacking the events in a time interval based on the exposure time. We also employ voxelization to obtain voxel representations. Specifically, given the original event stream $\mathcal{E}$ with range $H, W, T$, we divide the spatio-temporal 3D space into voxels with the size of each voxel being  $h', w', t'$. Hence, each voxel generally contains several events and the resulting event voxels in spatio-temporal space are of size $H/h', W/w', T/t'$. In practice, the above voxelization usually still produces tens of thousands of voxels. In order to further reduce the number of voxels and alleviate the effect of noisy voxels, we also adopt a voxel selection process to select top $K$ voxels based on the number of events contained in each voxel. 
Let $\mathcal{O}=\{o_1, o_2 \cdots o_K\}$ denote the collection of the final selected voxels. Each event voxel $o_i$ is associated with a feature descriptor $\textbf{a}_i\in \mathbb{R}^C$ which integrates the attributes (polarity) of its involved events. Hence, each $o_i\in \mathcal{O}$ is represented as: $o_i=(x_i,y_i,t_i, \textbf{a}_i)$, where $x_i,y_i,t_i$ denotes the 3D coordinate of each voxel.

\begin{figure*}
\center
\includegraphics[width=4.8in]{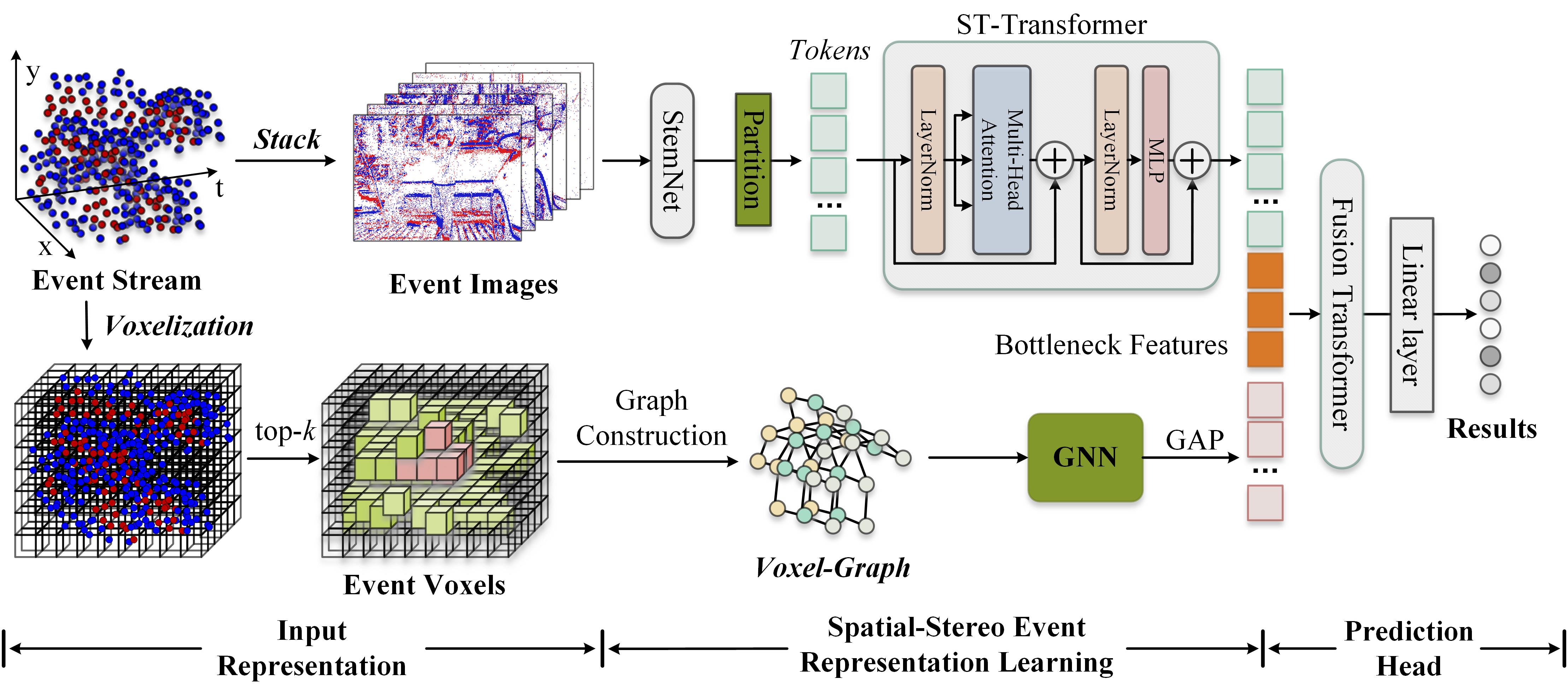}
\caption{An overview of our proposed Image-Voxel Feature Learning framework for event-based recognition.} 
\label{framework}
\end{figure*} 	

\noindent 
\textbf{Graph Neural Networks for Event Voxel Encoding. }  
We similarly construct a geometric neighboring graph $G^o(V^o, E^o)$ for voxel event data $\mathcal{O}$. To be specific, each node $v_i\in V^o$ represents a voxel $o_i=(x_i,y_i,t_i,\textbf{a}_i)\in \mathcal{O}$ which is described as a feature vector $\textbf{a}_i \in \mathbb{R}^{C}$. 
The edge $e_{ij}\in E^o$ exists between node $v_i$ and $v_j$, if the Euclidean distance between their 3D coordinates is less than a threshold R. We adapt Gaussian Mixture Model(GMM), convolution to learn the effective representations for voxel graph. To be specific, in each GCN layer, each event node $v_i$ aggregates the features from its adjacency nodes as 
\begin{align}
f'_d(v_i)\leftarrow \sigma\Big(\sum_{v\in V}\omega_{d}(v_i,v) f(v)\Big), d = 1, 2\cdots D
\end{align}
where $\sigma(\cdot)$ denotes the activation function, such as ReLU. $V$ denotes adjacency nodes of $v_i$. $\omega_{d}(v_i,v)$ denote the learnable convolution kernel weights. Finally, we adapt average graph pooling to get the global representation of voxel graph.

\noindent 
\textbf{Spatial-temporal Transformer for Event Frame Encoding. } 
After a series of data augmentation, each video sample obtained $T$ event frames with a size of ${H\times W}$. We extract initial CNN features and embed event frames through StemNet (ResNet18~\cite{he2016deep} is used in our experiments). After obtaining the initial features, we designed an ST-Transformer module to further achieve a better representation of spatio-temporal information. The proposed module consists of multi-head self-attention (MSA), MLP, and Layernorm (LN). As shown in  Fig.~\ref{framework}, $T$ event frames are divided into $N$ patches in spatial dimension, therefore, the ${T\times N}$ tokens can be obtained. We add learnable location encoding to these tokens and feed them into the ST-Transformer module to fully extract the enhanced spatio-temporal features, as shown in Eq.~\ref{STformer1} and Eq.~\ref{STformer2}:
\begin{align}
\label{STformer1}
& Y = {X}^{in} + MSA(LN({X}^{in}))\\
\label{STformer2}
& X^{out} = Y+MLP(LN(Y))
\end{align}

\noindent 
\textbf{Bottleneck Transformer. } 
In order to achieve the interaction between Event Images and Event Voxels information representations and learn a unified spatio-temporal context data representation. We also designed the Fusion Transformer module and introduced the Bottleneck mechanism. Specifically, let ${X}^{image}\in \mathbb{R}^{T\times N\times d}$ and ${X}^{voxel}\in \mathbb{R}^{1 \times d}$ represent the outputs of the previous ST Transformer and GNN modules, respectively. We first collect the ${T\times N}$ image and ${T\times N}$ randomly initialized Bottleneck tokens together and feed them to Fusion Transformer which includes multi-head self-attention (MSA) and MLP submodule, i.e., 
\begin{align}
& F^1 = [{X}^{image}, {X}^{bottleneck}] \in \mathbb{R}^{2\times T\times N\times d} \\
&\widetilde{F}^1 = FusionTransformer(F^1)
\end{align}
We then split  $\widetilde{F}^1$ into two parts, i.e., the images feature representation  $\widetilde{F}^{image}$ and the bottleneck feature representation $\widetilde{F}^{bottleneck}$. The latter one will be concatenated with ${X}^{voxel}$ and fed into the Fusion-Transformer module for interactive learning of the two representations. Similarly, 
\begin{align}
& F^2 = [{X}^{bottleneck},{X}^{voxel}] \in \mathbb{R}^{ (T\times N +1)\times d} \\
&\widetilde{F}^2 = FusionTransformer(F^2)
\end{align}
Finally, we concatenate both $\widetilde{F}^2$ and $\widetilde{F}^{image}$ together and flattened them into a feature representation. After that, we utilize a two-layer MLP to output the final class label prediction, as shown in Fig.~\ref{framework}. We adopt the Negative Log Likelihood Loss function~\cite{miranda2017understanding} to train the whole network.

\section{Experiment} 

\subsection{Dataset and Evaluation Metric} 

In this work, we utilized two datasets, namely DVS128-Gait-Day~\cite{wang2021eventgait3Dgraph}, N-MNIST~\cite{orchard2015nmnist}, and ASL-DVS~\cite{bi2020graph}, to evaluate our proposed model. Here is a brief introduction to these datasets: 

\noindent 
$\bullet$ \textbf{ASL-DVS}~\cite{bi2020graph}: This dataset consists of 100,800 samples, with 4,200 samples available for each letter. The focus was on the 24 letters representing the handshapes of American Sign Language. Each video in this dataset has a duration of approximately 100 milliseconds. The author captured these samples using an iniLabs DAVIS240c camera under realistic conditions. 

\noindent 
$\bullet$ \textbf{DVS128-Gait-Day}~\cite{wang2021eventgait3Dgraph} dataset is proposed for event-based gait recognition. It contains 4,000 videos corresponding to 20 classes. 20 volunteers are recruited for data collection using a DVS128 Dynamic Vision Sensor (the pixel resolution is $128 \times 128$). 

\noindent 
$\bullet$ \textbf{N-MNIST}~\cite{orchard2015nmnist} dataset is obtained by recording the display equipment when visualizing the original MNIST (28 × 28 pixels). The ATIS event camera is used for the data collection and each event sample lasts about 10ms. There are 70,000 event files for this dataset, the training and testing subset contains 60,000 and 10,000 videos, respectively. The resolution of this dataset is $28 \times $28. 

Note that the top-1 and top-5 accuracy are employed as the evaluation metrics throughout our study.

\subsection{Implementation Details}  

Our proposed dual-stream event-based recognition framework can be trained in an end-to-end manner. The initial learning rate is set as 0.001 and multiplied by 0.1 for every 60 epochs. We select eight frames for each video sample and divide each frame into eight tokens. For the constructed voxel graph, the threshold R is set to 2. The scale of the voxel grid is (4. 4. 4) for the ASL-DVS dataset. We select 512 voxels as the graph node for the structured graph representation learning. Our code is implemented using Python 3.8 and trained on a server with RTX3090 GPUs. 


\begin{table}
\center   
\caption{Results on the ASL-DVS~\cite{bi2020graph} dataset.} 
\label{ASLDVSResults}
\scalebox{0.8}{
\begin{tabular}{ccccccccccccccc} 		
\hline \toprule [0.5 pt] 
\textbf{EST}\cite{gehrig2019end}   &\textbf{AMAE}\cite{deng2020amae}     &\textbf{M-LSTM}\cite{cannici2020mlstm}    &\textbf{MVF-Net}\cite{deng2021mvfnet}     & \textbf{EventNet}\cite{sekikawa2019eventnet}\\  
0.979   & 0.984     &0.980     &0.971     &0.833    \\ 
\hline 
\textbf{RG-CNNs}\cite{bi2020graph}     &\textbf{\makecell[c]{EV-VGCNN}}\cite{deng2021evvgcnn}     &\textbf{\makecell[c]{VMV-GCN}}\cite{xie2022vmvgcn}     &\textbf{EV-Gait-3DGraph}\cite{wang2019evGait}   &\textbf{Ours} \\
0.901     &0.983     &0.989  &0.738  &0.996	 \\
\hline \toprule [0.5 pt] 
\end{tabular}
}
\end{table}

\subsection{Comparison with Other SOTA Algorithms}  
As shown in Table~\ref{ASLDVSResults}, previous works already achieve high performance on the ASL-DVS~\cite{bi2020graph} dataset. For example, the EST~\cite{gehrig2019end} (0.979), AMAE~\cite{deng2020amae}  (0.984), M-LSTM~\cite{cannici2020mlstm} (0.980), and MVF-Net \cite{deng2021mvfnet} (0.971). Note that the GCN-based model, VMV-GCN \cite{xie2022vmvgcn}, achieves better results, i.e., 0.989 on the top-1 accuracy. Thanks to the spatial-temporal feature learning and fusion network proposed in this work, we set new state-of-the-art performance on this dataset, i.e., 0.996. 
On the N-MNIST~\cite{orchard2015nmnist} dataset, as shown in Table~\ref{NMNISTResults}, we also achieve SOTA performance compared with recent strong models. These comparisons fully validated the effectiveness of our proposed framework for event-based recognition. We provide two figures to better illustrate our results, as shown in the left subfigure of  Fig.~\ref{ASLtop5TSNE}.

\begin{table*}
\center
\scriptsize    
\caption{Results on the N-MNIST~\cite{orchard2015nmnist} dataset.} 
\label{NMNISTResults} 
\begin{tabular}{ccccccccccccccccc} 		
\hline \toprule [0.5 pt] 
\textbf{\makecell[c]{EST}}\cite{gehrig2019end}   &\textbf{\makecell[c]{M-LSTM}}\cite{cannici2020mlstm} 
&\textbf{\makecell[c]{MVF-Net}}\cite{deng2021mvfnet}  &\textbf{\makecell[c]{Gabor-SNN}}\cite{sironi2018hats} 
&\textbf{EvS-S}\cite{li2021graph} \\
99.0   &98.6     &98.1     &83.7     &99.1 \\
\hline 
\textbf{HATS}\cite{sironi2018hats}    &\textbf{EventNet}\cite{sekikawa2019eventnet}     &\textbf{RG-CNNs}\cite{bi2020graph}   &\textbf{EV-VGCNN}\cite{deng2021evvgcnn}  &\textbf{Ours}  \\ 
99.1   &75.2     &99.0     & 99.4  &98.9	\\ 
\hline \toprule [0.5 pt] 
\end{tabular}
\end{table*}

\subsection{Ablation Study} 
To help researchers better understand the method we proposed, in this section, we conduct comprehensive experiments of component analysis on the DVS128-Gait-Day dataset and ASL-DVS dataset to check their influence on the overall model.

\begin{figure*}
\center
\includegraphics[width=\textwidth]{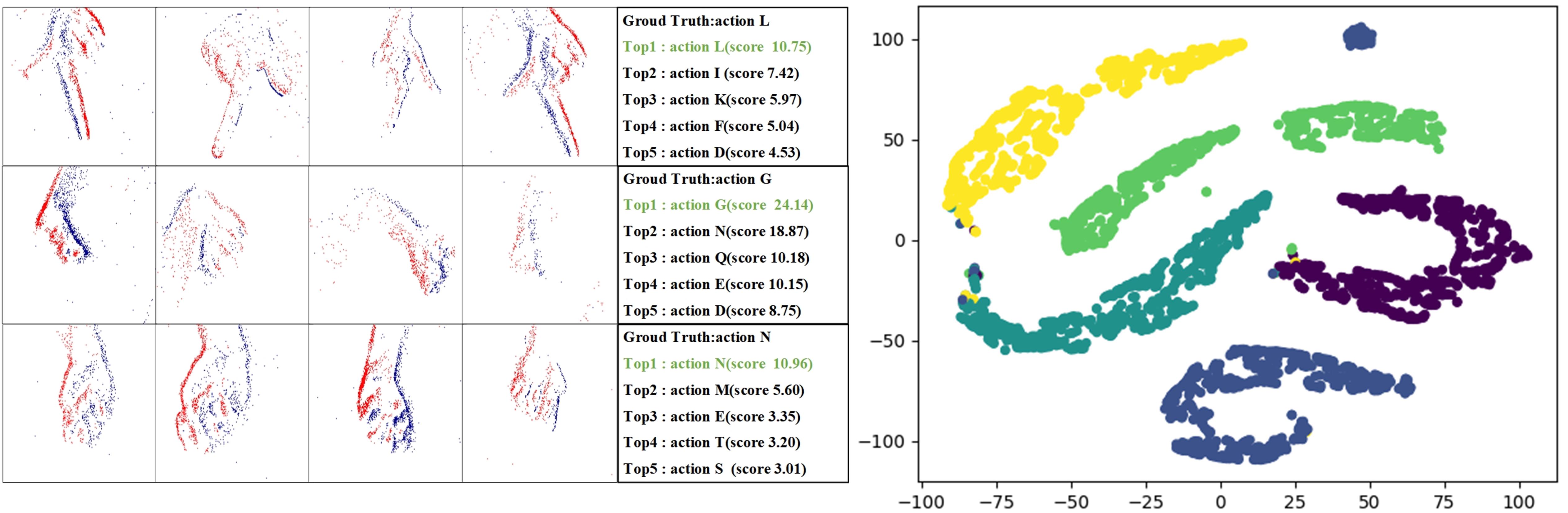}
\caption{Visualization of top-5 recognition results and feature distribution on the ASL-DVS dataset.}
\label{ASLtop5TSNE}
\end{figure*}

\textbf{Component Analysis. } 
Table~\ref{ASLtop5TSNE} shows the effect of using different components on experimental results. In this part, we didn't use the Bottleneck, and the dataset we use is DVS128-Gait-Day. 
\textbf{Event image only} indicates that we only transform the event flow into the synchronous event images, which gets the result of 95.2. \textbf{Event voxel only} indicates that we only employ voxelization to obtain the compressed event representation and it achieves 98.0. We also use Event image and Event voxel together when obtaining event representation, denoted by \textbf{Event Image}+\textbf{Voxel}. It gets the result of 98.7. We can easily draw the conclusion by comparing the above three cases that using Event Image and Event voxel together can achieve higher performance, which reflects the effectiveness of our method.

\noindent 
\textbf{Effect of Bottleneck.} 
In this paper, we use Bottleneck Transformer to enhance the performance when fusing the modules. As shown in Table~\ref{ASLResults}, \textbf{w/o Bottleneck Feature} means we do not feed the learning token into Bottleneck. It gets the result of 98.5. \textbf{w/o FusionFormer} means we do not use Fusion Transformer before the linear layer, in other words, we use the all components proposed in this paper except for the Bottleneck, and the result is 98.3. We found that compared with these two experiments, the result has increased after introducing the Bottleneck which indicates that the Bottleneck is a better choice for our framework. At the same time, a comparison with the experimental results in Table~\ref{ASLDVSResults} shows that the experimental result drops when there is no Bottleneck, which also demonstrates Bottleneck has a positive effect on our proposed model.


\begin{figure}
\begin{minipage}{.5\textwidth}
  \centering
  \begin{table}[H]
    \centering
    \caption{Ablation study on DVS128-Gait-Day dataset~\cite{wang2021eventgait3Dgraph}.} 
    \label{DVS128GaitResults}
    \scalebox{1}{
    \begin{tabular}{c|l|c} 		  
    \hline \toprule [0.5 pt] 
    Index   & Component    & Results   \\ 
    \hline
    1 &  Event image only             &  95.2        \\ 
    2 &  Event voxel only           &   98.0        \\ 
    3 &  Event Image + Voxel        & 98.7          \\ 
    \hline \toprule [0.5 pt] 
    \end{tabular}
    }
  \end{table}
\end{minipage}%
\hfill  
\begin{minipage}{.5\textwidth}
  \centering
  \begin{table}[H]
    \centering
    \caption{Ablation study on ASL-DVS~\cite{bi2020graph}.} 
    \label{ASLResults}
    \scalebox{1}{
    \begin{tabular}{c|l|c} 		  
    \hline \toprule [0.5 pt] 
    Index   & Component    & Results   \\ 
    \hline  
    1 &  w/o Bottleneck Feature     & 98.5  \\ 
    2 &  w/o FusionFormer           & 98.3          \\ 
    \hline \toprule [0.5 pt] 
    \end{tabular}}
  \end{table}
\end{minipage}
\end{figure}

\subsection{Parameter Analysis}  
The storage space of our proposed method is 220.3 MB. Our model spends 16.7 ms for each video in ASL-DVS dataset.

\section{Conclusion}     
Previous event-based recognition approaches typically represented event streams as point clouds, voxels, or images, and employed various deep neural networks to learn feature representations. However, these approaches are usually challenged by monotonous modal expressions and the design of the network structure. To overcome these challenges, this paper introduces a novel dual-stream framework for event representation, extraction, and fusion. The proposed framework simultaneously models two common representations: event images and event voxels. By leveraging Transformer and Structured Graph Neural Network (GNN) architectures, spatial information and three dimensional stereo information can be learned separately. Moreover, the introduction of a bottleneck Transformer facilitates the fusion of the dual-stream information. Extensive experiments were conducted to evaluate the performance of our framework, using two widely used event-based classification datasets. The results demonstrate that our proposed framework achieves state-of-the-art performance. These findings highlight the effectiveness of the dual-stream framework in addressing the limitations of existing approaches and improving the recognition accuracy in event-based object recognition tasks.

\noindent 
\textbf{Acknowledgement:} This work is supported by the National Natural Science Foundation of China (No. 62102205).

{ 
\bibliographystyle{IEEEtran}
\bibliography{reference}
}
\end{document}